\documentclass[runningheads]{llncs}
\usepackage{graphicx}
\usepackage{amsmath,amssymb} 
\usepackage{color}
\usepackage{array}
\usepackage{url}
\usepackage[width=122mm,left=12mm,paperwidth=146mm,height=193mm,top=12mm,paperheight=217mm]{geometry}

\definecolor{cmuColour}{RGB}{190,42,42}
\definecolor{gteaColour}{RGB}{42,190,42}
\definecolor{beoidColour}{RGB}{42,42,190}
\definecolor{beoidAMColour}{RGB}{255,211,0}
\definecolor{beoidASColour}{RGB}{246,18,97}
\definecolor{beoidAHColour}{RGB}{132,132,255}

\def\etal{\emph{et~al}}
\newcolumntype{C}[1]{>
{\centering\let\newline\\\arraybackslash\vspace{3pt}}m{#1}}

\begin{document}
\pagestyle{headings}
\mainmatter

\title{SEMBED: Semantic Embedding of Egocentric Action Videos} 

\titlerunning{SEMBED: Semantic Embedding of Egocentric Action Videos}

\authorrunning{Michael Wray, Davide Moltisanti,\\ Walterio Mayol-Cuevas and Dima Damen}

\author{Michael Wray\thanks{Both authors contributed equally to this work}, Davide Moltisanti$^\star$, Walterio Mayol-Cuevas and Dima Damen}


\institute{Department of Computer Science,\\
	University of Bristol, Bristol, UK\\
	\email{<FirstName>.<LastName>@bristol.ac.uk}
}

\maketitle

\begin{abstract}
We present SEMBED, an approach for embedding an egocentric object interaction video in a semantic-visual graph to estimate the probability distribution over its potential semantic labels. 
When object interactions are annotated using unbounded choice of verbs, we  embrace the wealth and ambiguity of these labels by capturing the semantic relationships as well as the visual similarities over motion and appearance features. We show how \mbox{SEMBED} can interpret a challenging dataset of 1225 freely annotated egocentric videos, outperforming SVM classification by more than 5\%.

\end{abstract}

\section{Introduction}
An egocentric camera captures rich and varied information of how the wearer interacts with their environment. 
The challenge for the visual understanding of this information is currently significant and not only incited by the enormous variety of such interactions but also by limitations in the available visual descriptors, e.g.~those rooted in motion or appearance.
Supervised learning from labelled examples is used to alleviate some of these ambiguities.
Egocentric datasets~\cite{Lee12,fathi2013modeling,spriggs2009temporal,Damen2014a} and interaction recognition methods~\cite{fathi2013modeling,Pirsiavash12,Fathi2012,McCandless13}
differ in the features used and classification techniques adopted, yet they all assume a semantically distinct set of \textit{pre-selected} verbs or verb-noun combinations for supervision. 
When free annotations are available - unbounded choice of verbs or verb-nouns - from audio scripts~\cite{Alayrac15learning} or textual annotations~\cite{Damen2014a}, a single label is selected to represent each interaction using a majority vote. Less frequent annotations are treated as outliers, though they typically represent a meaningful and correct annotation.
For example, lifting an object from a workspace could be described as \textit{pick-up}, \textit{lift}, \textit{take} or \textit{grab}; all valid labels.
Note that assuming multiple \textit{valid} labels is different from the problem of Ambiguous Label Learning, \cite{chen2015matrix,hullermeier2006learning}, where the aim is to find a single valid label from a mixed set of related and unrelated labels.

\begin{figure}[!t]
  \centering
    \includegraphics[width=1.0\textwidth]{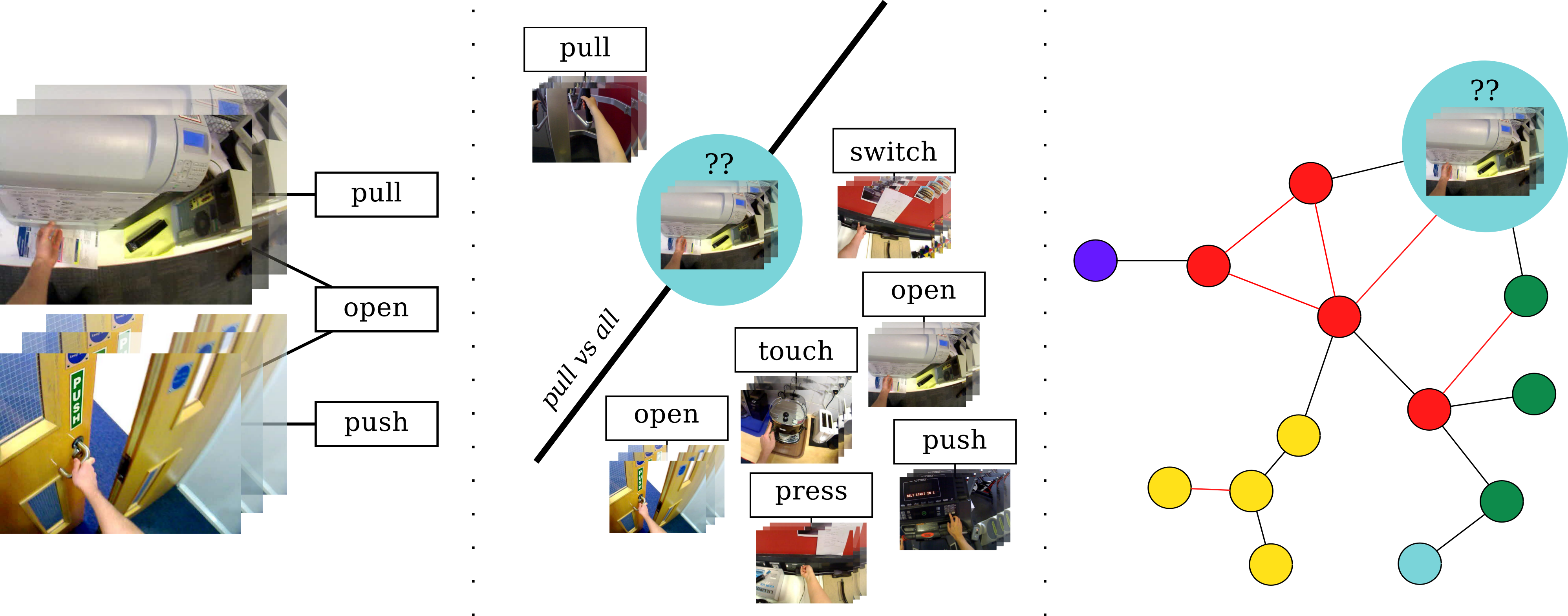}
    \caption{Given a dataset of free annotations, with potentially ambiguous semantic labelling~(left), 
    we propose to deviate from the one-vs-all classical approach~(middle) and instead build a graph that encapsulates semantic relationships and visual similarities in the training set~(right). Recognition then amounts to embedding an unlabelled video (denoted by `??') into the graph and estimating the probability distribution over potential labels.}
\label{fig:motivational}
\end{figure}

Egocentric video offers a unique insight into object interactions in particular.
The camera is ideally positioned to capture objects being used and, equally interesting, the different ways in which the same object is used.
One interaction (e.g.~\textit{open}) applies to a wide variety of objects, and each video can be labelled by multiple valid labels (e.g. \textit{open door} vs \textit{push door}).
In this context, recognition cannot be simplified as a one-vs-all classification task. Capturing the semantic relationships between annotations and the visual ambiguities between accompanying video segments can better represent the space of possible interactions.
Figure~\ref{fig:motivational} shows a graphical abstract of our work.

Given a dataset of egocentric object interactions with free annotations, we 
contribute four diversions from previous attempts:
(i)~We treat all free annotations as valid, correct labellings,
(ii)~A graph that combines semantic relationships with visual similarities is built, inspired by previous work on object class categories in single images~\cite{fang2012measuring}~(Sec.~\ref{sec:methodEmbed}),
(iii)~A~test video is embedded into the previously learnt semantic-visual graph and the probability distribution over its possible annotations is estimated (Sec.~\ref{sec:methodClass}) and
(iv)~When verb meanings are available, we discover semantic relationships between annotations using WordNet~(Sec.~\ref{sec:methodAH}). 

We test semantic embedding (SEMBED) on three public egocentric datasets \cite{Damen2014a,spriggs2009temporal,Fathi2012}. We show that as the number of verb annotations and their semantic ambiguities increase, SEMBED outperforms classification approaches. 
We also show that 
incorporating higher level semantic relationships, such as the hyponymy relationship, improves the results.
Note that while we focus on \textit{egocentric object interaction recognition} as a rich domain of semantic and visual ambiguities, some of the arguments can apply to action recognition in general.

\section{Embedding Object Interactions - Prior Work}
\label{sec:Ego_RW}

To the best of our knowledge, embedding for egocentric action recognition has not been attempted previously. We first review works on recognising egocentric
object interactions, then review works which incorporate semantic knowledge for recognition tasks.

\noindent \textbf{Egocentric Object Interaction Recognition:} \hspace{4pt}
Egocentric action recognition works range from self-motion~\cite{Kitani11} (e.g. walk, cycle) to  high-level activities (e.g.~\cite{spriggs2009temporal,kuehne2015towards,lade2010task,Bleser15,Sudeep10}).
On the task of object interaction recognition, approaches vary in whether they use hand-centred features~\cite{ishihararecognizing,kumarfly}, object-specific features~\cite{fathi2013modeling,Damen2014a,McCandless13,ren11} or a combination~\cite{Lee12,li2015delving}.
Ishihara~\etal~\cite{ishihararecognizing} use dense trajectories in addition to global hand shape features and apply a linear SVM to determine the action class. Kumar~\etal~\cite{kumarfly} sample and describe superpixel regions around the hand. Their method allows for hand detectors to be trained spontaneously with the user performing the action.

Object-specific features are better suited for recognising verb-noun actions (e.g. \textit{pick-cup vs pick-plate}) rather than a general \textit{picking} action.
In Damen \etal~\cite{Damen2014a}, spatio-temporal interest points have been used to discover object interactions in an unsupervised manner. 
The works of Fathi~\etal~\cite{fathi2013modeling,Fathi2012,li2015delving,fathi2011learning} have tested features including gaze, colour, texture and shape for verb-noun action classification.
Of these, \cite{fathi2013modeling} specifically discusses the change in the object state as a useful feature to recognise object interactions.
Though attempting video summarisation primarily, Ghosh~\etal~\cite{Lee12} introduces a collection of features that could be used to classify object-interactions such as distance from the hand, saliency, objectness represented using a spatio-temporal pyramid to detect change.
These features were proven useful for segmenting object-interactions from a lengthy video, but have not been tested for action classification \textit{per se}.
On several publicly available datasets, Li~\etal~\cite{li2015delving} compare motion, object, head motion and gaze information along with a linear SVM for object interaction classification.
Their results prove that Improved Dense Trajectories (IDT) proposed by~\cite{Wang2011} outperform other motion features.

With the emergence of highly-discriminative appearance-based features, pre-trained Convolutional Neural Networks (CNN) on ImageNet have also been tested.
In \cite{Moghimi_2014_CVPR_Workshops}, CNN is evaluated for distinguishing manipulation from non-manipulation actions on an RGB-D egocentric dataset.
Ryoo~\etal~\cite{Ryoo_2015_CVPR} combine CNN with IDT along with novel time series pooling for dog-centric manipulation and non-manipulation actions.
More recently, fine-tuned multi-stream CNN approaches have achieved state of the art results on egocentric datasets~\cite{Ma16,Singh16}, though are tuned on each dataset independently.

Based on~\cite{li2015delving,Ryoo_2015_CVPR} conclusions, in this work we report results on IDT as a state-of-the-art motion feature and pre-trained CNN features a state-of-the-art appearance feature. Testing tuned CNNs is left for future work.

\vspace*{8pt}
\noindent \textbf{Semantic Embedding for Object and Action Recognition:} \hspace{4pt}
Using linguistic semantic knowledge for Computer Vision tasks, including action recognition, has been fuelled by the accessibility of text or audio descriptions from online sources.

One such dataset which made this possible was gathered from {Y}ou{T}ube videos \cite{chen2011collecting} with free annotations. The dataset includes a variety of real-world scenarios, though not limited to egocentric or object-interactions.
For each video, multiple annotators were asked to describe the video.
 Both \cite{Motwani12,Guadarrama13} use this dataset for action recognition.
In Motwani and Mooney~\cite{Motwani12}, the most frequently annotated verb for each video is used, and verbs are grouped into classes using semantic similarity measures, extracted from the {W}ord{N}et hierarchy as well as information corpuses. 
Videos are described by HoG and HoF features around spatio-temporal interest points.
Guadarrama \etal~\cite{Guadarrama13} find subject, object and verb triplets in an attempt to automatically annotate the action. They create a separate semantic hierarchy for each, formulated by co-occurrences of words within the free annotations and use Spearman's rank to find the distances between  clusters. 
Semantic links are used to generate specific, rather than general, annotations and a classifier is trained for each leaf node within the hierarchies.
Their method allows zero-shot action annotation by trading-off specificity and semantic similarity.
While combining semantics, both works use majority voting to limit the description per class to a single verb.

Another recent {Y}ou{T}ube dataset was collected of users performing tasks while narrating their actions \cite{Alayrac15learning}.
Labels are extracted from audio descriptions using automatic speech recognition.
Verb labels are then used to align videos using a {W}ord{N}et similarity measure as well as visual similarity (HoF and CNN) to find the sequence of actions in a task. 

Semantics have also been used for object recognition with images.
Jin \etal \cite{jin2005image} use {W}ord{N}et to remove noisy labels from images which have multiple labels. 
Similarly, Ordonez \etal~\cite{ordonez2015predicting} use {W}ord{N}et to find the most frequently-used object labels amongst multiple annotations.
We build our work on Fang and Torresani \cite{fang2012measuring}, where images are embedded in a semantic-visual graph.
In~\cite{fang2012measuring}, images are clustered depending on the semantic relationships between the labels and edges of the graph are weighted with the visual similarity. They use {I}mage{N}et as the database for training, and benefit from the fact that images within {I}mage{N}et are organised according to the {W}ord{N}et hierarchy.
We differ from~\cite{fang2012measuring} in how we add visual links to the semantic graphs as will be explained next.

\section{Semantic Embedding of Egocentric Action Videos} 
\label{sec:method}
We next, in Sec.~\ref{sec:methodEmbed}, explain how we build a semantic-visual graph (SVG) that encodes label and visual ambiguities in the training set.
In Sec.~\ref{sec:methodClass}, we detail how videos with an unknown class are embedded in SVG, and how the probability distribution over their annotations is estimated.
Finally, in Sec.~\ref{sec:methodAH} we explore further semantic relationships when verb meanings are annotated.

  \subsection{Learning the Semantic-Visual Graph}
\label{sec:methodEmbed}

The Semantic-Visual Graph (SVG) is a representation of the training videos, with three sources of information encoded. First, videos that are semantically linked, e.g. have the same label, are linked in SVG. Second, nodes that are visually similar, yet semantically distinct, should also be linked as these indicate visual ambiguities.
Third, edge weights correspond to the normalised visual similarity, over neighbouring nodes, using a visual descriptor and a defined distance measure.
In this section we explain how SVG$_u$, an undirected graph, is constructed, then normalised to achieve the directed graph SVG.

SVG$_u$ is an undirected graph, where one node $x_i \in \textrm{SVG}_u$ corresponds to one training video.
Assume AX($x_i$, $x_j$) is a binary function that checks whether two video labels are semantically related.
Initially, AX($x_i$, $x_j$) is \textit{true} when both videos are annotated by the exact same verb.
This assumption is revisited in Sec.~\ref{sec:methodAH}.
Edges in SVG$_u$ are created between nodes with a semantic relationship:

\begin{equation}
x_i \frown x_j \in \textrm{SVG}_u \iff AX(x_i, x_j) = true
\end{equation}

\noindent The undirected edge $x_i \frown x_j \in \textrm{SVG}_u$ is assigned a weight $w_{x_i \frown x_j} = D_v (x_i, x_j)$ where $D_v$ is a distance measure defined over the visual descriptor chosen.
Assume $rank (D_v(x_i, x_j))$ is a function that returns the relative position of the distance measure amongst all the remaining pairs of videos such that, 
\begin{equation}
\resizebox{1.0\textwidth}{!}{$rank(D_v(x_i, x_j)) = n \iff D_v(x_i, x_j) = min_n(D_v(x_k, x_l)) \quad \forall x_k, x_l \in \textrm{SVG}_u \quad and \quad AX(x_k, x_l) \ne true$}
\end{equation}

\noindent and $min_n$ is the $n^{th}$ minimum element in the list. In addition, assume \\$rank_i (D_v(x_i, x_j))$ is a function that returns the relative position of $D_v(x_i,x_j)$ amongst all nodes not connected to $x_i$ such that,
\begin{equation}
\resizebox{1.0\textwidth}{!}{$rank_i(D_v(x_i, x_j)) = n \iff D_v(x_i, x_j) = min_n(D_v(x_i, x_l)) \quad \forall x_l \in \textrm{SVG}_u \quad and \quad AX(x_i, x_l) \ne true$}
\end{equation}

\noindent Further links are added to SVG$_u$ to encode visual ambiguities such that,
\begin{equation}
x_i \frown x_j \in \textrm{SVG}_u \iff rank(D_v(x_i, x_j)) \le m \quad or \quad rank_i(D_v(x_i, x_j)) = 1
\end{equation}

\noindent where $m$ is the number of visual connections in SVG$_u$ that correspond to the top $m$ visually similar and semantically dissimilar nodes in SVG$_u$.
We differ from~\cite{fang2012measuring} in that we ensure each node is connected to its top visually similar but semantically distinct node.

The undirected graph SVG$_u$ is then converted to a directed graph by replacing each edge with two directed edges.
\begin{equation}
x_i \frown x_j \in \textrm{SVG}_u \Rightarrow \{x_i \rightarrow x_j, x_j \rightarrow x_i\} \in \textrm{SVG}
\end{equation}

\noindent The weights of directed edges are initially the same as the weights for their undirected counterparts however they
are normalised to define the probability of traversing from video $x_i$ to $x_j$,
\begin{equation}
\label{eq:travProb}
P(x_i \rightarrow x_j)=\frac{1/w_{x_i\rightarrow x_j}}{\sum\limits_k{1/w_{x_i \rightarrow x_k}}} \quad \quad \forall x_i \rightarrow x_k \in \textrm{SVG}
\end{equation}

\noindent The reciprocal of the weights is taken so that the most visually similar path will have the highest probability.

\begin{figure}[t]
  \centering
    \includegraphics[width=1.0\textwidth]{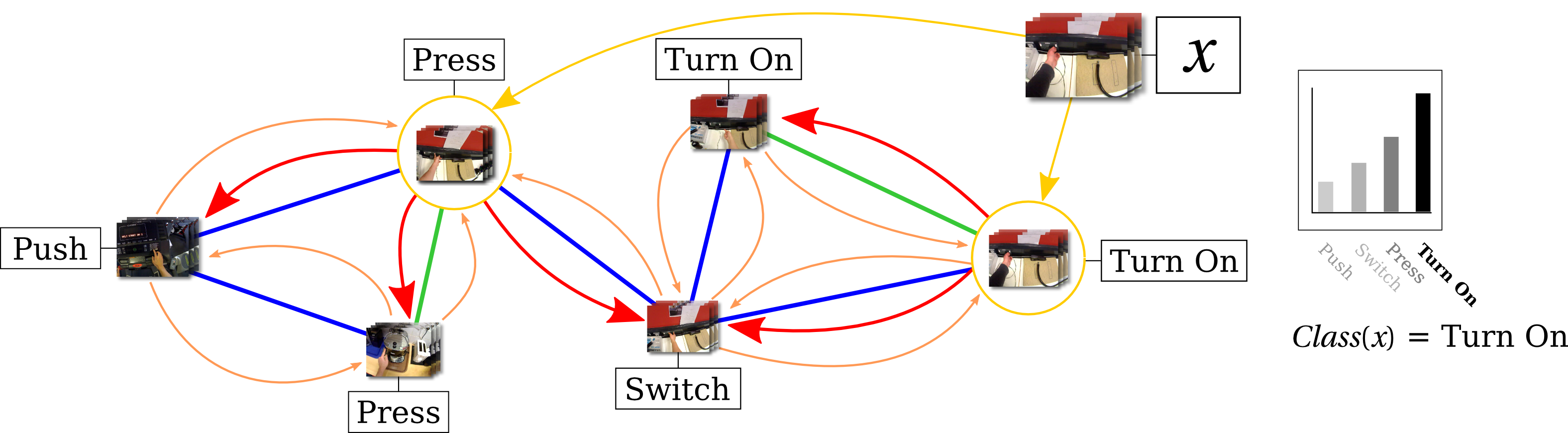}
    \label{eq:prob}
    \caption{The Semantic-Visual Graph (SVG) is built for training data, with  semantic links (green) and visual links (blue) between videos. Given a test video $x$, two nearest neighbours are found (yellow) and a Markov Walk of 2 steps (step1-red and step2-orange) finds the probability distribution over potential labellings. Ref. supplementary material for animation.}
\label{fig:graphEx}
\end{figure}

\subsection{Embedding in Semantic-Visual Graph}
\label{sec:methodClass}
Given a test video, $x$, we first embed the video into SVG then use the Markov Walk (MW) method from \cite{fang2012measuring} to determine $Class\,(x)$. 
To embed $x$, we begin by finding the set $\mathcal{R}$ which contains the $z$ closest neighbours to $x$ based on visual distance, such that
\begin{equation}
\mathcal{R} = \{x_i \in \textrm{SVG} \mid rank(D_v(x,x_j)) \le z\} 
\end{equation}

\noindent We embed $x$ into SVG by adding directed edges connecting $x$ to nodes in\\ \mbox{$\mathcal{R}$: $x \rightarrow x_i \quad \forall x_i \in \mathcal{R}$}
with normalised weights $P(x \rightarrow x_i)$.
Following the embedding, MW attempts to traverse the nodes in the directed graph to estimate the probability of $Class(x)$. Given the Markovian assumption and a predefined number of steps $t$, we calculate the probability distribution of reaching a node $x_i$ as follows
\begin{equation}
P(x_{i+t} \mid x) = \prod_{x_i \in R} \Bigl( P(x \rightarrow x_i) \prod_{j = 1}^t P(x_{i+j-1} \rightarrow x_{i+j}) \Bigr)
\end{equation}

\noindent To perform MW efficiently, we construct the vector q such that
\begin{equation}
 	q(i) = 
	\begin{cases}
     P(x \rightarrow x_i)  &x_i \in \mathcal{R} \\
     0 &\mbox{otherwise}
 	\end{cases}
     \label{eq:matQ}
 \end{equation}

\noindent We also construct a matrix $A$ such that $A(i,j)=P(x_i \rightarrow x_j)$ (Eq.~\ref{eq:travProb}), note that this matrix is asymmetrical as nodes have a different set of neighbours in SVG.
Accordingly, $P(x_{i+t} \mid x) = q^T A^{t}$
where $q^T$ is the transpose of $q$ and $t$ is the number of steps in MW.
We can then accumulate $P(Class(x))$ for every unique annotation $ax \in AX$ as follows

\begin{equation}
P(Class(x) = ax) = \sum_{AX(x_{i+t}, ax) = true} P(x_{i+t} \mid x)
\end{equation}

\noindent We then select $\arg\max_{Class(x)}P(Class(x))$ as the semantic label of $x$.
Figure~\ref{fig:graphEx} shows an example of SVG and video embedding. In the figure, given two nearest neighbours $z = 2$ and two steps in MW $t = 2$, the probability distribution over possible labellings is calculated.

\subsection{Semantic Relationships: Synsets and Hyponyms}
In Sec.~\ref{sec:methodEmbed}, videos are considered semantically linked only when the annotated verbs are the same.
SVG then enables handling ambiguities via incorporating visual similarity links in the graph.
However, further semantic relationships, such as synonymy and hyponymy relationships, can be exploited between annotations.
In linguistics, two words are \textit{synonyms} if they have the same meaning, and the set of all synonyms is a \textit{synset}.
Moreover, two words are described as a \textit{hyponym} and a \textit{hypernym} respectively if the first is a more specific instance of the second.
The terms originate from the Greek word $hyp\acute{o}$ and $hyp\acute{e}r$ - \textit{under} and \textit{over}.

\begin{figure}[b!]
  \centering
   \includegraphics[width=1.0\textwidth]{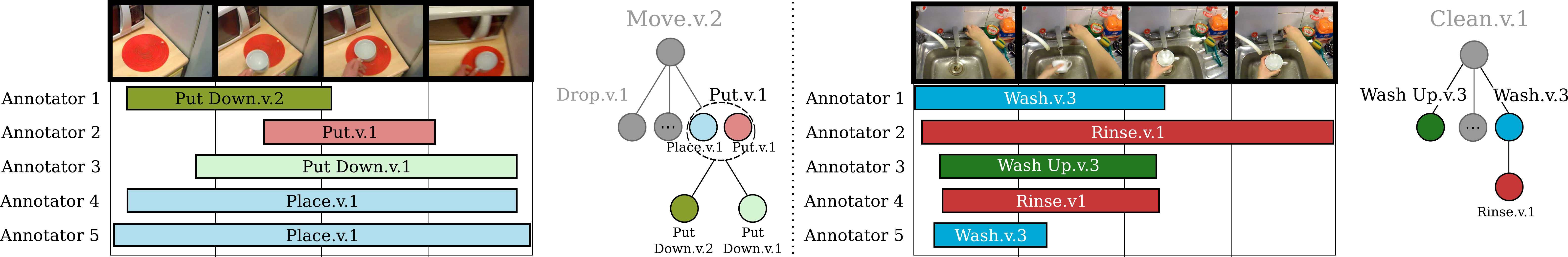}
   	\caption{
    Five free annotations for two sequences from the BEOID dataset~\cite{Damen2014a}, and the respective semantic relationships between the annotations from WordNet~\cite{miller1995wordnet}. In the hierarchy, each parent-child relationship represents a hypernym-hyponym pair. The dotted circle encapsulates a synonymy relationship. The start and end times of the actions are also shown. For placing a cup on a mat (left), synonyms $put.v.1$ and $place.v.1$ were chosen by annotators. $put\_down.v.1$, a hyponym of $put.v.1$ was also used. For washing a cup (right), the verbs $wash.v.3$, $wash\; up.v.3$ and $rinse.v.1$ were chosen. $rinse.v.1$ is a hyponym of $wash.v.3$.} 
\label{fig:annotationEx}
\end{figure}

Synonymy and hyponymy relationships are encoded in lexical databases. WordNet (v3.1, 2012) is a commonly-used lexical database that
is based on six semantic relations~\cite{miller1995wordnet}.
In the WordNet verb hierarchy, verbs are first separated into their various meanings by the notation $\langle word \rangle.v.\langle s\rangle$ where $s \ge 1$ is the number of disjoint meanings. 
The meanings are then arranged in hierarchies that encapsulate semantic relationships.
To benefit from such hierarchies, verbs should be annotated with their meanings.
We annotate~\cite{Damen2014a} using verb meanings, and Fig.~\ref{fig:annotationEx} shows how such annotations of the same action can be synonyms and hyponyms, as annotators chose different or more specific action descriptions. 
\label{sec:methodAH}

Given annotated meanings, we define the term action synsets (AS) to indicate that annotations are linked by a synonymy relationship solely, and the term action hyponym (AH) to indicate that annotations are linked by both the synonymy or the hyponymy relationships.
For comparison, we define the term action meaning (AM) where annotations are linked only when the annotation matches exactly.
We use the general term AX where $AX \in \{$AM, AS, AH$\}$ is one of the the possible types of semantic relationships tested.

\section{Datasets, Experiments and Results}
\label{sec:datasets}
\begin{figure}[t]
\centering
    \begin{minipage}[b]{0.8\textwidth}
    
        \resizebox{\textwidth}{!}{
            \centering
            \begin{tabular}{|l|c|c|c|c|c|}
                \hline
                \textbf{Name} & \textbf{Users} & \textbf{Seq.} & \textbf{OI Seg.s} & \textbf{Used OI Seg.s} & \textbf{Semantic Verbs}  \\ \hline
                \textcolor{cmuColour}{CMU} ~\cite{de2008guide} & 5 & 35 & 516 & 406 & 12 (33.8, 30.5) \\ \hline
                \textcolor{gteaColour}{GTEA+} ~\cite{Fathi2012}& 13& 30 & 3371 & 1000$^{\star}$ & 25 (40.0, 75.5) \\ \hline
                \textcolor{beoidColour}{BEOID} ~\cite{Damen2014a} & 3-5 & 58 & 1488 & 1225 & 75 (16.3 34.2) \\ \hline
            \end{tabular}
        }
        \centering
        \vspace*{6pt}
        \includegraphics[width=0.7\textwidth]{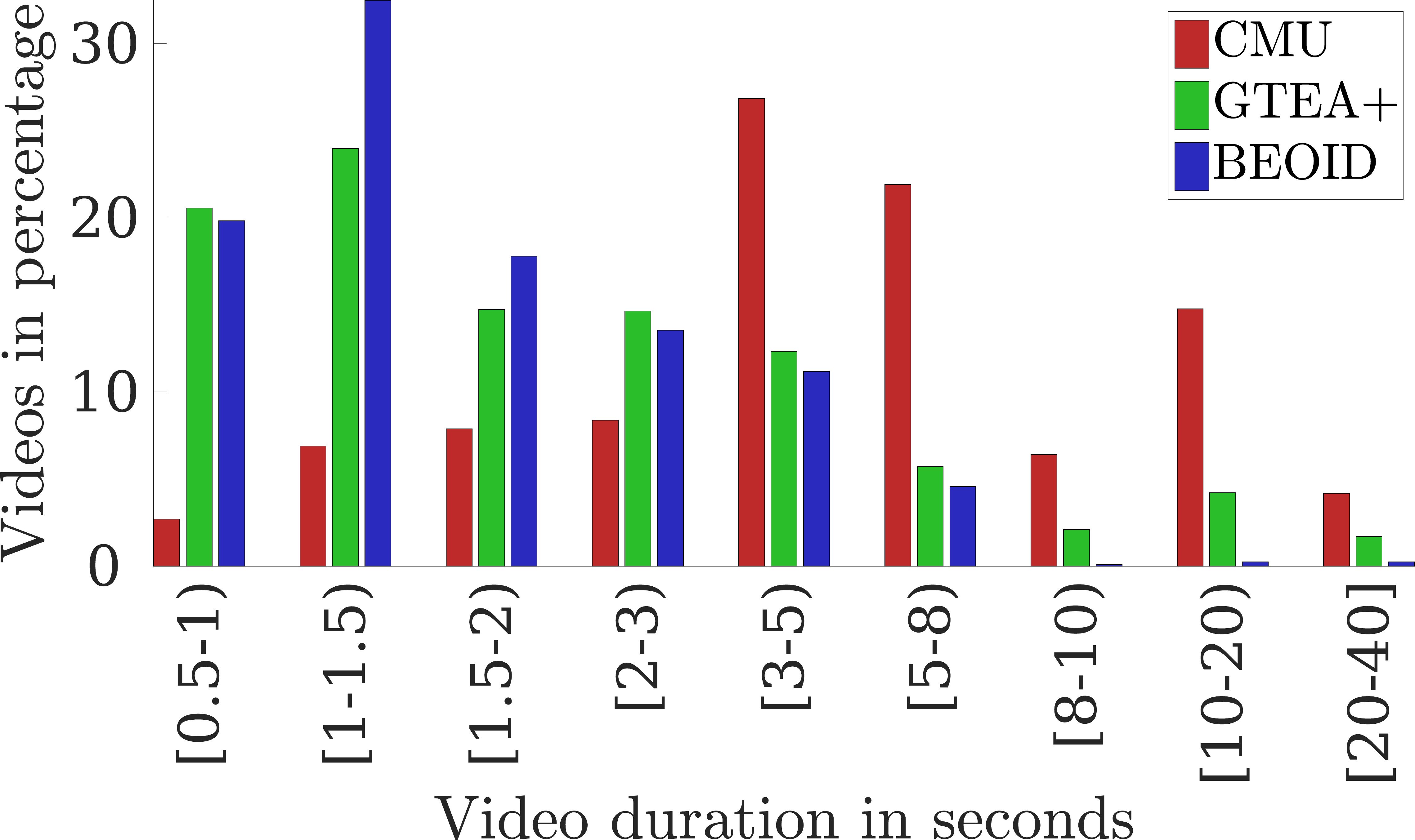}
    \end{minipage}

    \caption{Dataset details (top) and video length distributions (bottom). Number of users, segments, Object-Interaction (OI) segments and used segments in the results (length $<$ 40s) are detailed. We report the number of annotated verbs along with $\mu$ and $\sigma$ for the number of segments per verb. $^{\star}$: Due to the size of GTEA+ we sampled 1000 videos randomly. Ref. supplementary material for frequencies of verb annotations per dataset.}
    \label{fig:datasetInfo}
\end{figure}

We selected three publicly available datasets that primarily focus on object interactions from egocentric videos~\cite{de2008guide,Fathi2012,Damen2014a} (Figure \ref{fig:datasetInfo}).

\noindent \textbf{Verb annotations:} We exploited the annotations provided by the authors to split the CMU and GTEA+ sequences into object-interaction segments. For CMU, object-interaction annotations are only provided for the activity of \textit{making brownies}.
Annotators chose from 12 disjoint verbs to ground-truth segments. 
In GTEA+ annotators chose from verb-noun pairing to  ground-truth, e.g. \textit{cut\_{}cucumber} versus \textit{divide\_{}bun} and similarly \textit{squeeze\_{}ketchup} versus \textit{compress\_{}bun}.
When removing the nouns, verbs could be used interchangeably but free annotations were not available to annotators.

While BEOID contains a variety of activities and locations, ranging from a desktop to operating a gym machine, it does not provide action-level annotations so we annotated BEOID using free annotations\footnote{Annotations can be found at: \url{http://www.cs.bris.ac.uk/~damen/BEOID/}}, allowing annotators to split video sequences into object-interaction segments in addition to choosing the verb.
We recruited 20 native English speakers.
These annotators were given a free textbox to label each segment with the verb that best described the seen interaction \textit{in their opinion}. Once
a verb has been chosen, the annotators were given the set of potential meanings extracted from WordNet for the chosen verb. Again, they were asked to select the meaning that, \textit{in their opinion}, best suited the segment. 
Multiple annotators (8-10) were asked to label each task to intentionally introduce variability in the choice of verbs and start-end times of object interaction segments.

\noindent \textbf{Motion and Appearance Features:} We test two state-of-the-art feature descriptors to represent both the motion and the appearance of the videos.
These are the Improved Dense Trajectories~(IDT)~\cite{Wang2013} and Overfeat Convolutional Neural Networks pre-trained for ImageNet classes~(CNN)~\cite{sermanet-iclr-14}. 
For CNN features, we take every 5th frame from 30fps video, starting always from the first frame, and rescale to 320x240 pixels.

\noindent \textbf{Encodings:} We test two encodings, using Bag of Words (BoW) \cite{csurka2004visual} and Fisher Vectors (FV)~\cite{Sanchez2013} with Euclidean distance. For IDT, when creating the BoW and FV representations, we use a 25\% random sample from every video to model the Gaussians for efficiency.
We vary the number of Gaussians ($\gamma_{fv}$) and the size of the codebook ($\gamma_{bow}$) in reported results.

\noindent \textbf{Classification:}
In all results, leave-one-person-out cross validation has been used. Namely, when testing a video containing one person performing an action, all other videos captured from the same person are excluded from the training set.
For SVM results, as the tested datasets contain an imbalance in the distribution of instances per class, we weight the classes by the term $w(c) = 1/prior(c)^\lambda$
where $\lambda \in [0,1]$ is the exponent that best fits the distribution of segments per verb for a given dataset (ref supplementary material).

\begin{table}[t]
    \centering
    \caption{As the number of verbs increases from \textcolor{cmuColour}{12} to \textcolor{beoidColour}{75}, the best performance changes from SVM to SEMBED. Results are obtained with $\gamma_{fv} = 10$ and $\gamma_{bow} = 256$, $k=$\{\textcolor{cmuColour}{3},\textcolor{gteaColour}{5},\textcolor{beoidColour}{5}\}, $m$ = 240, $z$=\{\textcolor{cmuColour}{2},\textcolor{gteaColour}{6},\textcolor{beoidColour}{4}\}, $t$=\{\textcolor{cmuColour}{20},\textcolor{gteaColour}{20},\textcolor{beoidColour}{8}\} for CNN and $z$=\{\textcolor{cmuColour}{4},\textcolor{gteaColour}{5},\textcolor{beoidColour}{14}\}, $t$=\{\textcolor{cmuColour}{4},\textcolor{gteaColour}{20},\textcolor{beoidColour}{10}\} for IDT. For completion, state-of-the-art results on verb-noun classes are reported under `Other Works' thus are not directly comparable to our verb only results.} 
    \resizebox{\textwidth}{!}{%
        \begin{tabular}{|l|C{1.0cm}|C{1.3cm}|C{2.0cm}||C{1.0cm}|C{1.3cm}|C{2.0cm}||C{1.0cm}|C{1.3cm}|C{2.0cm}||C{1.0cm}|C{1.3cm}|C{2.0cm}||C{1.0cm}|C{2.6cm}|}
            \cline{1-13}
            \textit{FEATURES} & \multicolumn{6}{c||}{\textbf{CNN}} & \multicolumn{6}{c||}{\textbf{IDT}} &\multicolumn{2}{c}{}\\ \cline{1-13}
            \textit{ENCODING} & \multicolumn{3}{c||}{\textbf{FV}}  & \multicolumn{3}{c||}{\textbf{BOW}} & \multicolumn{3}{c||}{\textbf{FV}}& \multicolumn{3}{c||}{\textbf{BOW}} &\multicolumn{2}{c}{} \\ \hline
            \textit{METHOD}   & \textbf{\footnotesize{SVM}} & \textbf{\footnotesize{K-NN}} & \textbf{\footnotesize{SEMBED}} & \textbf{\footnotesize{SVM}} & \textbf{\footnotesize{K-NN}} & \textbf{\footnotesize{SEMBED}} & \textbf{\footnotesize{SVM}} & \textbf{\footnotesize{K-NN}} & \textbf{\footnotesize{SEMBED}} & \textbf{\footnotesize{SVM}} & \textbf{\footnotesize{K-NN}} & \textbf{\footnotesize{SEMBED}} & \textbf{Verbs} &\textbf{Other Works} \\ \hline

            \textcolor{cmuColour}{\textbf{CMU}}~\cite{de2008guide} & 58.6 & 46.6 & 46.3 & 55.9 & 43.3 & 52.0 & \textcolor{cmuColour}{\textbf{69.4}} & 58.1 & 57.4 & 55.9 & 57.6 & 61.6 &\textcolor{cmuColour}{\textbf{12}} & 48.6 \cite{spriggs2009temporal}, 73.4 \cite{taralova2011source} \\ \hline
            \textcolor{gteaColour}{\textbf{GTEA+}}\cite{Fathi2012} & 15.6 & 30.0 & 31.0 & 25.1 & 33.5 & 33.6 & \textcolor{gteaColour}{\textbf{43.6}} & \textcolor{gteaColour}{\textbf{43.4}} & 42.1 & 27.8 & 34.5 & 40.3 & \textcolor{gteaColour}{\textbf{25}} &60.5 \cite{li2015delving}, 65.1~\cite{Ma16} \\ \hline
            \textcolor{beoidColour}{\textbf{BEOID}}~\cite{Damen2014a} & 20.9 & 34.4 & 37.5 & 15.2 & 19.1 & 19.6 & 38.7 & 36.0 & 37.4
            & 34.8 & 39.6 & \textcolor{beoidColour}{\textbf{45.0
            }} & \textcolor{beoidColour}{\textbf{75}} &- \\ \hline
        \end{tabular}%
    }
    \label{table:resultsTableAV}
\end{table}

\begin{figure}[!h]
    \centering
    \includegraphics[width=1.0\textwidth]{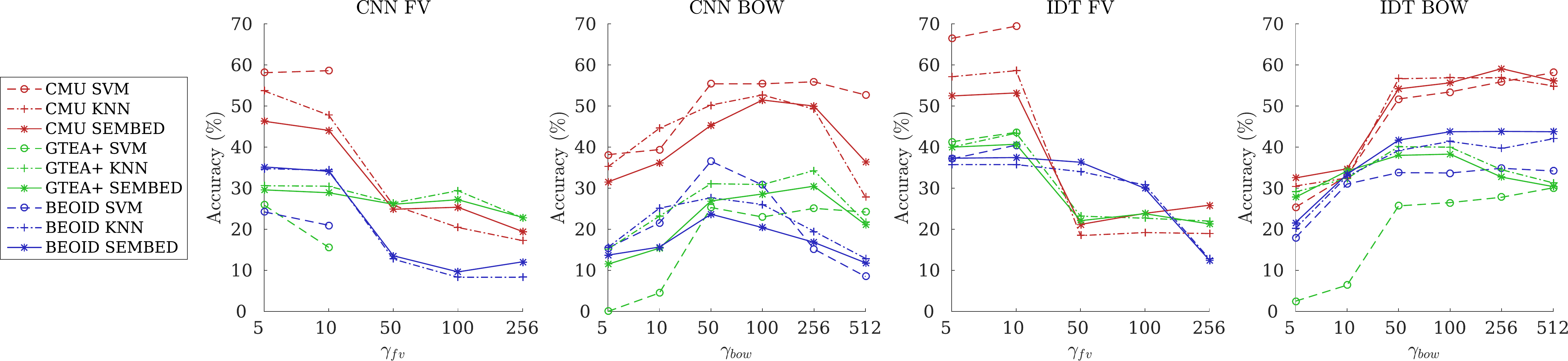}
    \caption{Results as $\gamma_{fv}$ and $\gamma{bow}$ vary for CMU, GTEA+, BEOID. Results were shown with $k=5$, $m = 240$, $z=10$, $t=10$. Similar performance is seen for other parameters.}
    \label{fig:gammaFigure}
\end{figure}

\noindent\textbf{Results on annotated verbs:} \hspace{4pt}
\label{sec:ResultsAV}
Table~\ref{table:resultsTableAV} compares the three datasets for every $\langle$features, encoding, classifier$\rangle$ combination.
The following conclusions can be made: (i) for all datasets, motion features~(IDT) outperform appearance features~(CNN) when classifying verbs without considering the object used. (ii) for CMU and GTEA+, we produce comparable results to published results using motion information on the same datasets. These are reported under `Other Works' but are not directly comparable as published works tend to report on verb-noun classes.
(iii) For the three datasets with varying number of verbs, as the number of verbs increases (\textcolor{cmuColour}{12} $\rightarrow$ \textcolor{beoidColour}{75}) with an increase in semantic ambiguity, SEMBED outperforms standard classifiers (SVM and K-NN). While the table shows the best results for encoding, Fig.~\ref{fig:gammaFigure} reports comparative results as $\gamma$ is changed - $\gamma_{fv}=10$ generally led to higher accuracies on all datasets, compared to $\gamma_{bow} = 256$.

\begin{figure}[b]
    \includegraphics[width=1\textwidth]{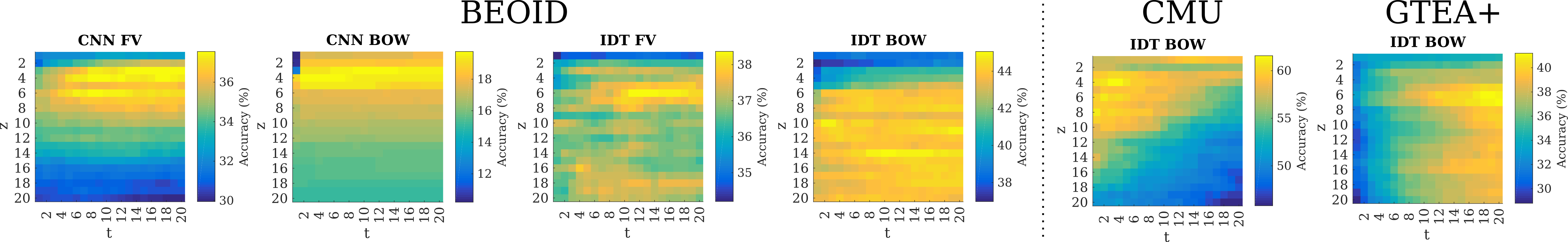} 
    \caption{Evaluation of SEMBED sensitivity to $z$ and $t$ parameters with $m = 240$.}
    \label{fig:sembedZandT}
\end{figure}

We test the sensitivity of SEMBED to its key parameters $z$ and $t$ and report results in
Fig.~\ref{fig:sembedZandT} showing the accuracy over various features for BEOID and across the three datasets for IDT-BOW (Ref. supplementary material for all combinations).
As noted, $z$ and $t$ behave differently for the various appearance and motion descriptors as well as for different encodings.
Generally, SEMBED is more sensitive to the choice of $z$ than $t$.
This is because the Markovian Walk (MW) is unable to represent the probability distribution over labels unless the starting positions are representative of the visual ambiguity.
Figure~\ref{fig:sembedZandT} also shows that MW isn't too helpful for CMU (as $t$ increases, accuracy decreases) because it
has visually distinctive verb classes.
On all datasets, SEMBED is resilient to changing $m$ values; the results are comparable on $180 \le m \le 400$.

 \noindent\textbf{Results on annotated verbs and meanings:} \hspace{4pt}
As mentioned earlier, we also annotate BEOID with verb-meaning ground-truth. This resulted in 108 $\langle word \rangle.v.\langle s\rangle$ annotations for the 1225 segments in the dataset. 
Note the increase in the number of classes from 75 when using verbs only to 108 when using verb-meaning ground-truth.
This increase is due to two reasons - one \textit{helpful}, another \textit{problematic}.
For example, it is \textit{helpful} when annotators choose between $hold.v.1$: \textit{``keep in a certain state, position''} and $hold.v.2$: \textit{``hold in one's hand''}.
Annotators would then use the first for when a button is pressed and the second for when an object is grasped.
However, frequently, WordNet meanings can appear ambiguous resulting in \textit{problematic} cases, especially in the context of egocentric actions. An example of this is the action of turning a tap on so water would flow. Annotators used $turn.v.1$: \textit{``change orientation or direction''} and $turn.v.4$: \textit{``cause to move around or rotate''} interchangeably. 
In WordNet though, $turn.v.1$ and $turn.v.4$ are not semantically related, introducing unwanted ambiguity affecting the ground-truth labels. While we accept that WordNet may not be the best method to incorporate meaning, we report results as semantic links are incorporated.

\begin{table}[t]
    \centering
    \caption{As synonymy (AS) and then hyponymy (AH) semantic relationships are incorporated, accuracy increases for all features on the BEOID dataset. $\gamma_{fv} = 10$, $\gamma_{bow} = 256$, $m$ = 240,  \{AM,AS,AH\}: $z_{CNN}$=\{3,3,2\}, $t_{CNN}$=\{20,20,14\}, $z_{IDT}$=\{6,10,13\}, $t_{IDT}$=\{20,20,2\}.}
    \resizebox{\textwidth}{!}{%
        \begin{tabular}{|l|C{1.0cm}|C{1.3cm}|C{2.0cm}||C{1.0cm}|C{1.3cm}|C{2.0cm}||C{1.0cm}|C{1.3cm}|C{2.0cm}||C{1.0cm}|C{1.3cm}|C{2.0cm}||C{1.5cm}|}
            \cline{1-13}
            \textit{FEATURES} & \multicolumn{6}{c||}{\textbf{CNN}} & \multicolumn{6}{c||}{\textbf{IDT}} \\ \cline{1-13} \textit{ENCODING} & \multicolumn{3}{c||}{\textbf{FV}} & \multicolumn{3}{c||}{\textbf{BOW}} & \multicolumn{3}{c||}{\textbf{FV}} & \multicolumn{3}{c||}{\textbf{BOW}} \\ \hline
            \textit{METHOD}   & \textbf{\footnotesize{SVM}} & \textbf{\footnotesize{K-NN}} & \textbf{\footnotesize{SEMBED}} & \textbf{\footnotesize{SVM}} & \textbf{\footnotesize{K-NN}} & \textbf{\footnotesize{SEMBED}} & \textbf{\footnotesize{SVM}} & \textbf{\footnotesize{K-NN}} & \textbf{\footnotesize{SEMBED}} & \textbf{\footnotesize{SVM}} & \textbf{\footnotesize{K-NN}} & \textbf{\footnotesize{SEMBED}} & \textbf{Classes} \\ \hline

            AM 		& 13.2 & 24.6 & 26.2 & 12.1 & 7.8 & 11.7 & 25.9 & 28.5 & 32.2 & 26.1 & 31.6 & \textbf{38.2} &108\\ \hline
            AS 		& 17.9 & 25.6 & 27.1 & 12.7 & 8.1 & 12.7 & 29.8 & 30.4 & 33.5 & 29.6 & 33.6 & \textbf{40.6} &102\\ \hline
            AH 		& 18.1 & 25.0 & 26.9 & 12.2 & 7.4 & 16.3 & 36.2 & 33.1 & 34.5 & 29.1 & 35.2 & \textbf{41.9} &84\\ \hline
        \end{tabular}%
    }
    \label{table:resultsTableASH}
\end{table}

\begin{figure}[t]
    \label{fig:sembedExample}
    \includegraphics[width=1.0\textwidth]{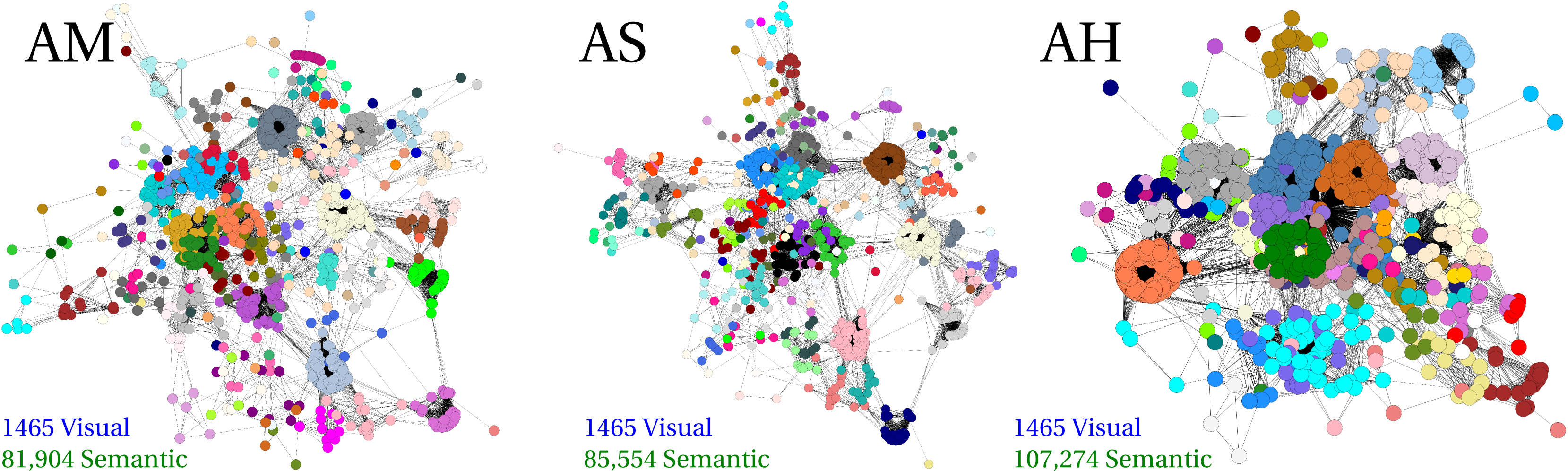} 
    \includegraphics[width=1.0\textwidth]{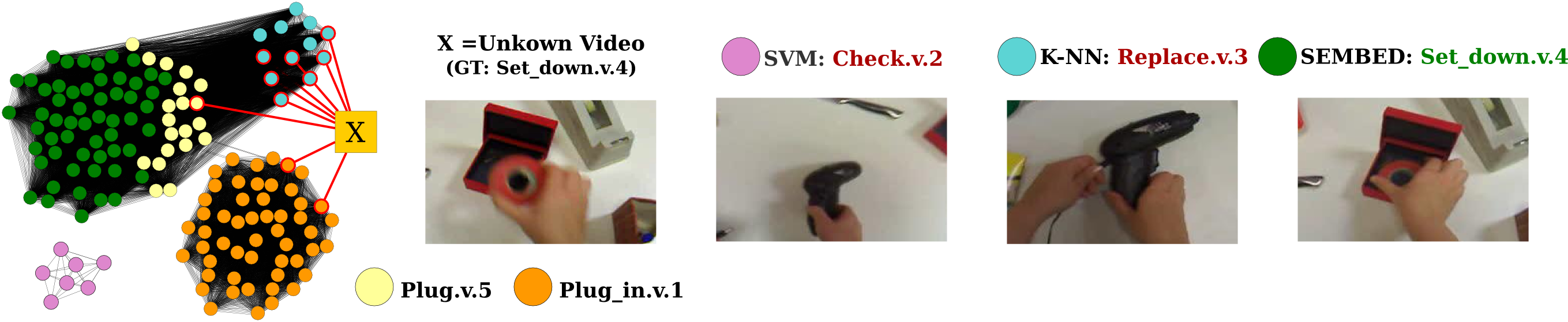} 
    \caption{SVG for three semantic levels on BEOID (top). Example using AH (bottom), SVM and K-NN produce incorrect results. The Markov walk of SEMBED allows the video to be correctly classified.}
    \label{fig:ex_sembed}
\end{figure}
We test the three types of semantic relationships $AX = \{$AM,AS,AH$\}$. 
Histograms of all classes for the various semantic relationships are included in the supplementary material.
Table~\ref{table:resultsTableASH} 
shows that embedding consistently improved performance as synsets and then hypernyms are grouped. Results also demonstrate the advantages of introducing semantic links between videos. Additionally, IDT continues to outperform CNN. 
Figure~\ref{fig:ex_sembed} shows one example of SEMBED in action when using meanings and AH semantic links\footnote{Video with results available at: \url{youtube}}.
It should be noted that the best performance of SEMBED on meanings is inferior to using verbs only. This is due to the difficulty in assigning meanings to verbs as previously noted.
Approaches to address meaning ambiguities are left for future work.

\section{Conclusion and Future Directions}
The paper proposes embedding an egocentric action video in a semantic-visual graph to estimate the probability distribution over potentially ambiguous labels.
SEMBED profits from semantic knowledge to capture interchangeable labels for the same action, along with similarities in visual descriptors.

While showing clear potential, outperforming classification approaches on a challenging dataset, results merely evaluate the $\arg\max$ label when compared to ground-truth.
Further analysis of the probability distribution will be targeted next.
Other approaches to identify semantically related object-interaction labels from, for example, other lexical sources, overlapping annotations or object labels will also be attempted. 
SEMBED's ability to scale to other object interactions and more discriminative visual descriptors will also be tested.

\bibliographystyle{splncs03}

\end{document}